\documentclass[lettersize,journal]{IEEEtran}
\usepackage{amsmath,amsfonts}
\usepackage{amsmath, amssymb, bm}
\usepackage{algorithmic}
\usepackage{algorithm}
\usepackage{array}
\usepackage{textcomp}
\usepackage{stfloats}
\usepackage{url}
\usepackage{verbatim}
\usepackage{graphicx}
\usepackage{cite}
\usepackage{subfigure}
\usepackage{tabularx}
\begin{document}

\title{Direct Covariance-to-Beamformer Mapping for Low-Latency Hybrid ISAC Beamforming with Subsequent DoA Estimation} 

 \author{Anjali K.S.$^{1}$, Manoj Kumar Panda$^{1}$, Chandramani Singh$^{2}$\\~\IEEEmembership{$^{1}$Department of Electronics and Communication Engineering, \\Amrita School of Engineering, Bengaluru, Amrita Vishwa Vidyapeetham, India\\ $^{2}$Department of Electronic Systems Engineering, Indian Institute of Science, Bangalore, India}}

\maketitle

\begin{abstract}
Integrated sensing and communication (ISAC) systems typically employ a two-stage beamforming pipeline, in which target directions are first estimated from the received covariance matrix and then used to compute the beamforming weights by solving non-convex optimization problems. Although effective, this approach introduces processing latency that can limit beam adaptation in dynamic environments. This letter proposes a beamforming deep neural network (BFDNN) for direct covariance-to-beamformer learning in ISAC systems. BFDNN is a compact regression network that maps covariance-matrix features directly to the hybrid beamforming weights produced by a hybrid ISAC beamformer, eliminating the direction-of-arrival (DoA) estimation stage during inference. Compared with MUSIC, classification-DNN, and regression-DNN based beamforming pipelines, BFDNN generates comparable beam patterns and higher effective spectral efficiencies while reducing the beamforming prediction latency by 25-30 times on the evaluated platform, demonstrating the potential of direct covariance-to-beamformer learning for low-latency ISAC beam updates. Moreover, the target directions can be inferred indirectly from the generated radiation pattern, yielding DoA accuracy competitive with that of dedicated learning-based methods in the evaluated scenario. 
\end{abstract}

\begin{IEEEkeywords}
ISAC, Dual-function radar communication (DFRC), DoA estimation, Hybrid beam forming, End-to-end beamforming, Deep learning, MUSIC, SVD, Communication, Radar, linear frequency modulation (LFM).
\end{IEEEkeywords}

\section{Introduction}

\IEEEPARstart{I}{SAC} has emerged as a relevant research technology for future wireless systems that enables radar sensing and wireless communication functionalities to share spectrum, hardware, and signaling resources\cite{ref6}. Applications such as connected vehicles, autonomous driving, unmanned aerial systems, and beyond-5G/6G networks require simultaneous environmental awareness and reliable wireless connectivity, making fast and accurate beamforming a fundamental component of ISAC operation. In particular, highly directional transmission is required to  steer energy toward targets for detection and toward users for communication, over limited spectrum resources. 

Most existing ISAC beamforming frameworks employ a two-stage processing pipeline. The received signal covariance matrix is first used to estimate target directions through DoA estimation algorithms such as MUSIC \cite{ref56} or learning-based DoA estimators \cite{ref2,ref3,ref4}. Based on the estimated directions, beamforming weights are computed using optimisation-based or decomposition-based beamformer design techniques \cite{ref1}. While this approach can provide accurate beam steering, its computational cost increases with antenna-array size and requires additional processing stages during online operation. Accordingly, beamforming latency becomes a concern in dynamic ISAC scenarios where rapid beam adaptation is required. 

Recent deep-learning approaches have improved DoA estimation accuracy and robustness under challenging operating conditions\cite{ref50,ref51}. However, most of these methods still retain the  DoA-estimation stage prior to beamformer computation. Other learning-based beamforming approaches often rely on long channel-state-information histories  \cite{ref40,ref42,ref62} or large feature representations\cite{ref64}, resulting in increased model complexity and training requirements\cite{ref41,ref63}. 

Motivated by these observations, we investigate whether hybrid beamforming weights\cite{ref1} can be predicted directly from the received covariance matrix without explicitly estimating DoAs. Since the covariance matrix already captures the spatial structure of the received signals, it provides a compact representation from which beamforming weights may be learned directly. This formulation transforms beamforming into a covariance-to-beamformer mapping problem and removes the need for conventional two-stage beamforming pipeline. 

To this end, this letter proposes a Beamforming Deep Neural Network (BFDNN) for both sine and LFM-based ISAC systems\cite{ref21}. The proposed network uses covariance matrix features as input and directly predicts hybrid beamforming weights through a simple  regression architecture. Unlike conventional approaches that separately perform DoA estimation and beamformer design, BFDNN generates beamforming weights through a single forward pass of the network. The overall processing concept is illustrated in Fig. \ref{fig_3}. Although the proposed framework does not explicitly estimate DoAs, target directions may be recovered from the peaks of the generated radiation pattern when required. This provides an additional sensing capability without introducing a dedicated DoA-estimation stage during inference.

The main contributions of this letter are summarized as follows:
\begin{itemize}
\item{A novel covariance-to-beamformer learning framework is proposed for hybrid ISAC beamforming,  that generates beamforming weights directly from covariance-matrix features.}
\item{ A compact regression-based neural-network model, termed BFDNN, is developed to eliminate separate DoA estimation stage during beamforming weight inference.}
\item{The framework is evaluated for  both sine and LFM signals}
\item{For an LFM-based ISAC scenario, the simulation results demonstrate that the proposed approach generates comparable  beam patterns and higher effective spectral efficiencies compared to benchmark methods.}
\item{The proposed framework reduces beamforming prediction latency from $3.3\text{–}3.9\times10^{-3}$ s to $1.3\times10^{-4}$s, approximately 25–30 times compared to benchmark methods on the evaluated platform.} 
\item{ Target directions are indirectly estimated from the radiation pattern generated from the proposed framework and result demonstrate competitive DoA-estimation accuracy to benchmark methods.}
\end{itemize}
The remainder of this letter presents the proposed covariance-to-beamformer framework, followed by performance evaluation and concluding remarks.

 \section{PROPOSED COVARIANCE-TO-BEAMFORMER FRAMEWORK}
 The proposed framework directly predicts hybrid beamforming weights from the received signal covariance matrix, eliminating explicit DoA estimation stage during inference. The overall concept is illustrated in Fig.\ref{fig_3}, where in the proposed processing chain, covariance-matrix features are mapped directly to beamforming weights through a lightweight neural network, unlike conventional approaches that first estimate DoAs and subsequently compute beamforming weights. 
\subsection{Signal Model and Covariance Matrix}
\begin{figure}
\centering
\includegraphics[width=\columnwidth]{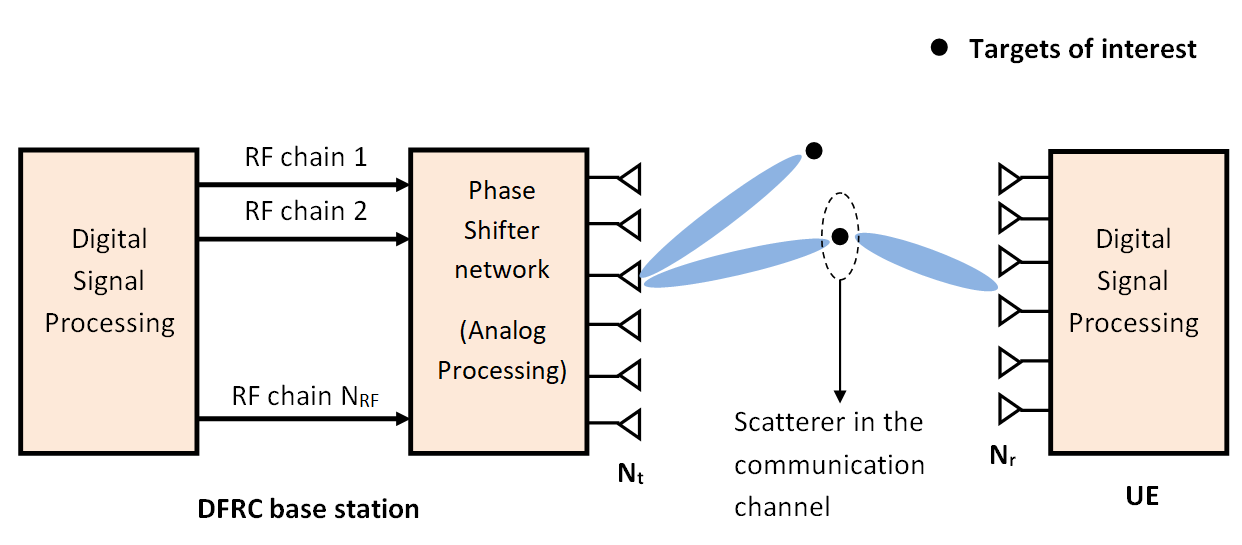}
\caption{Hybrid beamforming ISAC Scenario }
\label{fig_2}
\end{figure}

\begin{figure}[t]
\centering
\includegraphics[width=\columnwidth]{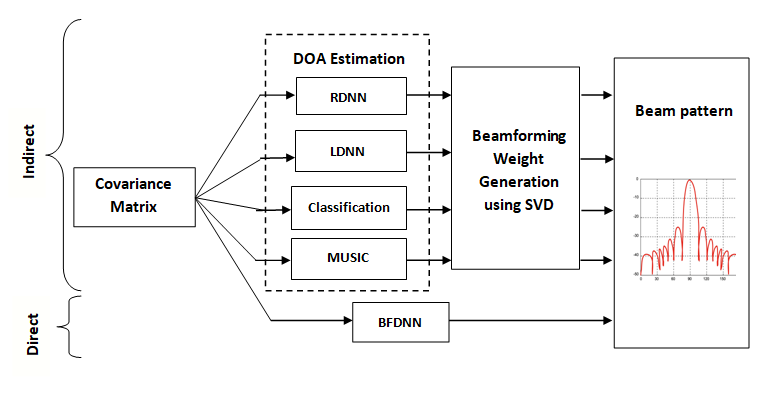} 

\caption{Different approaches analyzed in this paper}

\label{fig_3}
\end{figure}

An ISAC/ DFRC base station (BS) is considered, as shown in Fig. \ref{fig_2}, which performs target detection and downlink communication simultaneously, with the target detected via the communication scatterer. The BS adopts the hybrid analog-digital (HAD) beamforming framework of \cite{ref1} and is equipped with a uniform linear array (ULA) of $N_t$ antenna elements and $N_{\mathrm{RF}} \leq N_t$ RF chains, while the user equipment (UE) employs a ULA of $N_r$ antenna elements. Following \cite{ref1}, the BS transmits omnidirectional pilot signals and receives the echoes from $K$ targets, so that the received signal at snapshot $t$, is expressed as
\begin{equation}
\label{eqn_1}
    \mathbf{y}(t) = \sum_{k=1}^{K} \mathbf{a}(\theta_k) s_k(t) + \mathbf{e}(t) = \mathbf{A}(\theta) \mathbf{s}(t) + \mathbf{n}(t), \quad t = 1, \dots, T.
\end{equation}

Here, $\mathbf{A}(\theta)$ is the steering matrix of the $N_{t}$ element ULA with dimension of  $N_{t} \times K$ and  $\theta_k$ is the DoA of targets. The $\mathbf{s}(t) $  contains the reflected signal components, while $\mathbf{n}(t)$ represents additive white gaussian noise added to the signals.

$\mathbf{a}(\theta_k)$ is the $N_{t}$ dimensional steering vector, described as:

\begin{equation}
    \mathbf{a}(\theta_k) = \begin{bmatrix}
        1,
        e^{j 2 \pi \frac{d}{\lambda} \sin(\theta_k)},  
        \dots,  
        e^{j 2 \pi \frac{d}{\lambda} \sin(\theta_k)(N_t-1)}
    \end{bmatrix}.
\end{equation}

Here, $d$ is the distance between elements of the matrix and $\lambda = \frac{c}{f}$ is the wavelength at carrier frequency $f$, where $c$ is the speed of light.
 The covariance matrix is estimated from T received snapshots as
\begin{equation}
    \mathbf{R}_y = \mathbb{E}[\mathbf{y}(t) \mathbf{y}^H(t)] = \mathbf{A}(\theta) \mathbf{R}_s \mathbf{A}^H(\theta) + \sigma_{N_t}^2 \mathbf{I}_{N_{t}}
\end{equation}

The  $ \mathbf R_s=\mathbb{E}[\mathbf{s}(t)[\mathbf{s}^H(t)]$, $\sigma_{N_{t}}^2 $ is the noise variance, and $I_{N_t}$ is an identity matrix with a size of $N_t\times N_t$.
In reality, the matrix in (2) is unknown and is replaced by its sample estimate, 
\begin{equation}
    \tilde{\mathbf{R}}_y = \frac{1}{T} \sum_{t=1}^{T} \mathbf{y}(t) \mathbf{y}^H(t),
\end{equation}
This $\tilde{\mathbf{R}}_y$ is used for testing.

\subsection{Covariance Feature Representation}
The covariance matrix contains the spatial information required for beamforming-weight prediction. The feature here is a real-valued feature vector which contains the upper triangular elements of the covariance matrix excluding the diagonal elements. As diagonal elements contain noise variance effect and lower triangular elements are conjugate copies of upper triangular elements, those elements are removed from feature vector.
Let,
\begin{equation}
\\\mathbf{b}=\left[\mathrm{R}_{1,2}, \mathrm{R}_{1,3}, \cdots, \mathrm{R}_{1, M}, \mathrm{R}_{2,3}, \cdots, \mathrm{R}_{2, M}, \cdots, \mathrm{R}_{M-1, M}\right]^{\mathrm{T}}\
\end{equation}
denote the vector formed from the upper-triangular covariance entries where, $\mathrm{R}_{i,j}$ is the $i-th$ row $j-th$ column element in $\mathbf{R}_y$ matrix. The input feature vector is then constructed in such a way that real, imaginary and phase of these selected elements are included in it as:

\begin{gather*}\mathbf{z}=\left[{Real}\left\{\mathbf{b}^{\mathrm{T}}\right\}, {Imag}\left\{\mathbf{b}^{\mathrm{T}}\right\},{Phase}\left\{\mathbf{b}^{\mathrm{T}}\right\}\right]^{\mathrm{T}} /\Vert\mathbf{b}\Vert_2 \tag{6}\label{z_eqn}\end{gather*}

This representation preserves the amplitude and phase relationships contained in the covariance matrix while reducing redundancy arising from unnecessary elements.
\subsection{Beamforming Deep Neural Network}
\begin{figure}[t]
\centering
\includegraphics[width=\columnwidth]{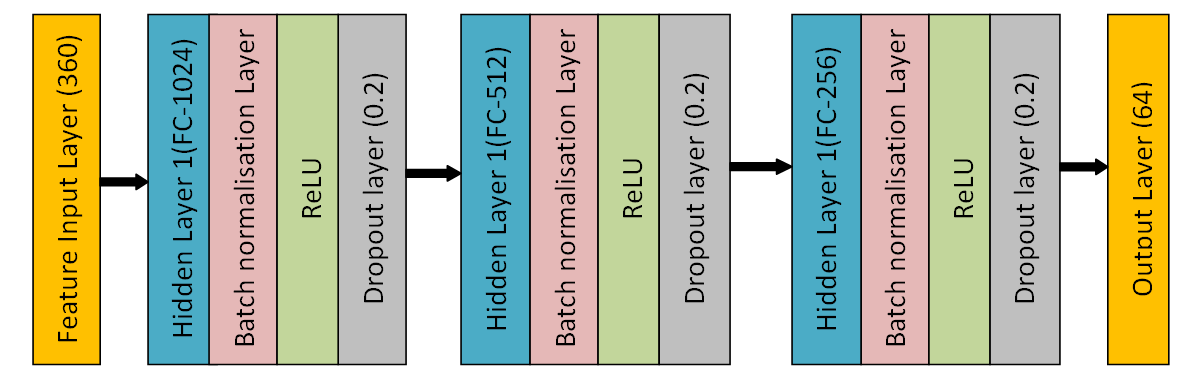}
\caption{BFDNN Architecture }
\label{fig_4}
\end{figure}

The proposed BFDNN learns a direct covariance-to-beamformer mapping,  $\mathbf z \rightarrow \mathbf w$,  where $\mathbf z$ is the covariance feature vector and $\mathbf w$ denotes the hybrid beamforming weights. The network architecture is shown in Fig. \ref{fig_4}. 

The architecture of the BFDNN was finalized through systematic experimentation. Different network configurations were tested based on performance to map from 360 input features to 64 output weights. The architecture follows a funnel-shaped architecture with layer dimensions $360\rightarrow 512 \rightarrow 256 \rightarrow 128 \rightarrow 64$ which progressively transform covariance features into a compact beamforming representation. The output layer contains 64 neurons corresponding to the real and imaginary components of the predicted complex beamforming coefficients. BFDNN consists of an input layer followed by three fully connected hidden layers and a linear output layer. Hidden layers employ batch normalisation, rectified linear unit (ReLU) activation, and dropout regularisation.

Compared with conventional DoA-estimation-based pipelines, BFDNN produces beamforming weights through a single forward pass of the network, thereby avoiding online DoA estimation and subsequent beamformer computation. 

\subsection{Training Objective}

The network is trained using supervised learning. Training labels are generated using the hybrid beamforming design of [1], which provides the target beamforming solution for each covariance-matrix input.
The  custom loss function used here for training this architecture is 
\begin{gather*}
L =   \frac{1}{N} \sum_{i=1}^{N} \left| R_s - \hat R_{\text{s}} \right|
\tag{34}\end{gather*}
where $R_{s}$ and $\hat R_{\text{s}}$ is denote the target and predicted transmit covariance matrices, respectively. $N$ denotes the number of elements in $R_{s}$ matrix. By minimizing the difference between transmit covariance matrices, the network is encouraged to generate beamforming weights that produce similar spatial transmission characteristics. Once trained, BFDNN directly predicts beamforming weights from covariance features without requiring explicit DoA estimation, angular search, or iterative optimization during inference.

\subsection{Training Dataset}
A supervised learning approach is adopted, in which the sources are assumed to lie within $[G_{\min}, G_{\max}]$ and their angular spacing $\Delta$ is drawn from $\{\Delta_{\min}, \Delta_{\min}+\Delta_d, \ldots, \Delta_{\max}\}$, as in \cite{ref3, ref4}, so that both closely spaced and widely separated sources are covered. Since the covariance-based input is independent of the DOA ordering, the first DOA $\theta_1$ is sampled on a grid $\eta$ from $\theta_{\min}$ to $\theta_{\max}-(K-1)\Delta$, and the remaining DOAs are $\theta_k=\theta_1+(k-1)\Delta$, $k=2,\ldots,K$. Even though, test SNR is unknown, the BFDNN is trained at a single SNR of 5 dB, yielding $118+116+\cdots+60=2670$ training instances. The inputs are the covariance-derived vectors of (6), and the label for each input is the real part followed by the imaginary part of the HAD beamforming weights generated using steps involving non convex optimization and Singular Value Decomposition (SVD) as in \cite{ref1}.
\section{PERFORMANCE EVALUATION}
The proposed BFDNN is evaluated in the hybrid beamforming ISAC framework of \cite{ref1} with simple sine and LFM signals. Simulations consider a base station equipped with a $N_t=16$ element ULA operating at 2.4 GHz, uniformly separated by a half-wavelength, i.e., $(d=\lambda/2)$. Two targets are assumed $(K=L=2)$, and  signals with $T=1000$ snapshots are used for covariance estimation. LFM signals are with bandwidth of 100 kHz . The proposed approach is compared with other benchmark approaches provided in Table.I. All the methods are implemented using MATLAB in Windows operating system running on the 12th Gen Intel(R) Core(TM) i9-12900H processor at 2.5 GHz and with an NVIDIA GeForce RTX 3080 Ti Laptop GPU.
\begin{table*}[h]
\label{tab:models}
\centering
\caption{Method-Specific Configuration of DL models}
\begin{tabularx}{\linewidth}{l X c X c X}
\hline
\textbf{Method} & \textbf{Input Representation} & \textbf{Input Size} & \textbf{Output Representation} & \textbf{Output Size} & \textbf{Training SNRs} \\
\hline

Proposed BFDNN &
 Upper triangular covariance matrix (Real, imaginary, and phase) &
360 &
Beamforming weights (real and imaginary parts) &
64 &

5dB \\

Classification Method \cite{ref2} &
Covariance matrix  (real, imaginary, phase) &
$16 \times 16 \times 3$ &
Binary angle grid vector &
121 &
$ \{-20,-15,-10,-5,0\} $dB \\
RDNN \cite{ref3}&
Upper triangular covariance matrix (Real and imaginary) &
240 &
Continuous DoA estimates &
2 &
$\{-13, -10, -5, 0\} $ dB \\

LDNN \cite{ref4}&
Average of upper triangular covariance matrix elements along super-diagonals (Real and imaginary) &
30 &
Continuous DoA estimates &
2 &
$ \{-13, -10, -5, 0\} $ dB \\

\hline
\end{tabularx}
\end{table*}
\subsection{Simulation Settings}
For training, we set $G_{\min}=-60^\circ$, $G_{\max}=60^\circ$, $\Delta_{\min}=2^\circ$, $\Delta_{\max}=60^\circ$, $\Delta_d=2^\circ$, and $\eta=1^\circ$. The benchmark methods use their best-performing configurations (mini-batch size of 32, gradient threshold of 1, and L2 regularization of $10^{-3}$) and are trained on the SNR ranges reported in their original works to ensure faithful reproduction of their best performance. The BFDNN follows the architecture in Fig.~\ref{fig_4}, with an input of size $3\times\frac{N_t(N_t-1)}{2}=360$ (real, imaginary, and phase parts of the upper-triangular covariance elements) and an output of 64 neurons, corresponding to the real and imaginary parts of the 32 complex HAD beamforming weights. All layers use `He' initialization, a gradient threshold of 8, L2 regularization of $10^{-4}$, and a mini-batch size of 36. All DL models are trained using the Adam optimizer with an initial learning rate of $10^{-3}$, reduced by a factor of 0.7 under a piecewise schedule every 10 epochs, for at most 500 epochs, with validation every 100 iterations and early stopping at a patience of 10. All methods are evaluated under identical test conditions with targets located at $\theta_1=-10.11^\circ$ and $\theta_2=26.11^\circ$.

\subsection{Indirect Direction Estimation from Beam Patterns}

\begin{figure}[!t]
\centering
\includegraphics[width=3.0in]{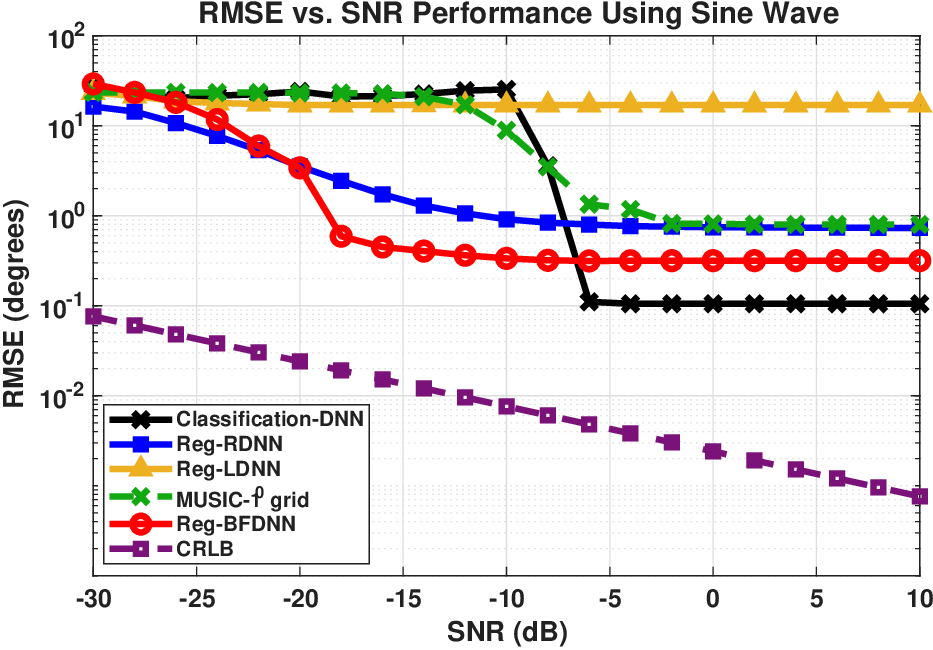}
\caption{RMSE vs SNR when the signals are sine waves.}
\label{fig_5}
\end{figure}

\begin{figure}[!t]
\centering
\includegraphics[width=3.0in,height=2.1in]
{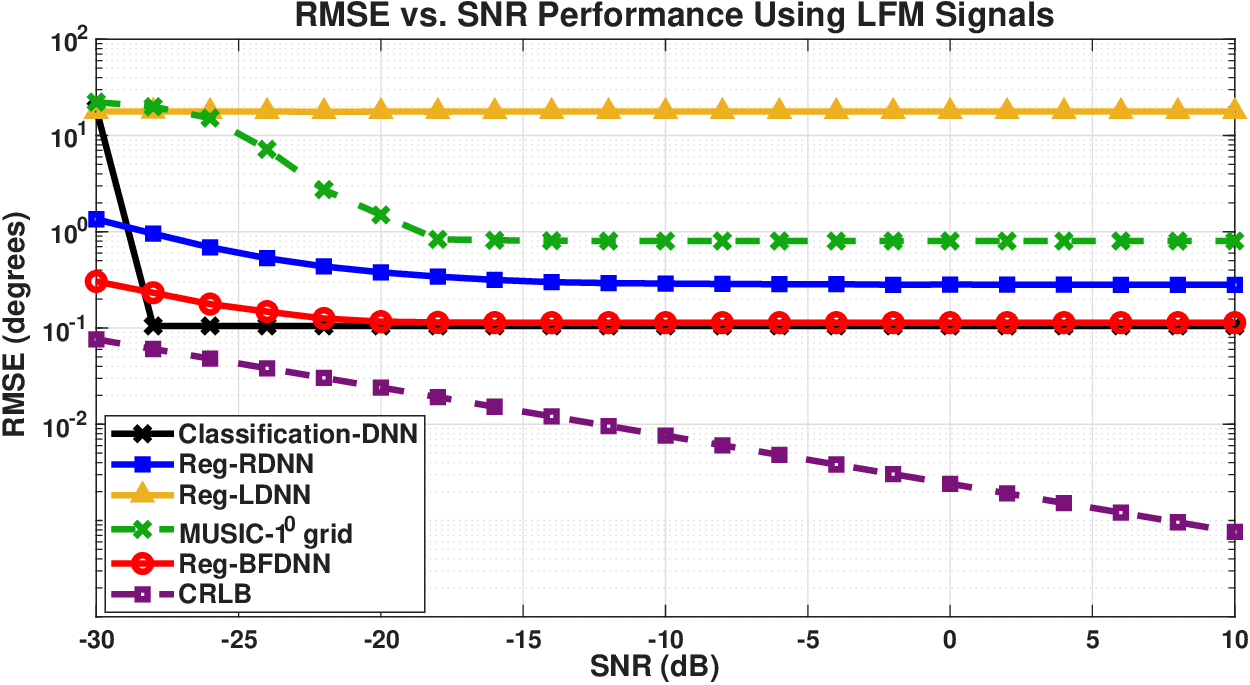}
\caption{RMSE vs SNR when the signals are LFM waves.}
\label{fig_6}
\end{figure}
The proposed BFDNN is designed to predict beamforming weights rather than target directions. However, target directions can be indirectly estimated from the resulting radiation pattern by locating the dominant beam peaks. To assess this capability, the angular locations of the principal lobes are used to estimate target directions and the corresponding root-mean-square error (RMSE) is compared with dedicated DoA-estimation approaches.

The DoA estimation accuracy of all the methods are compared in Fig. \ref{fig_5} when basic sine waves are considered as signals. Here  classification-DNN approach showcase good performance when compared to all other methods at higher SNRs from -6dB. Our proposed method shows the very next best approach compared to other methods.The classification method demonstrates high RMSE at low SNR levels (below $-10$ dB), where noise dominates and reduces feature separability. As the SNR increases beyond $-6$ dB, a sharp drop in RMSE is observed, indicating a transition from noise-limited to signal-dominant conditions. This abrupt change reflects the thresholding behavior of classification models, wherein performance improves significantly once the input features become sufficiently distinguishable.

Fig. \ref{fig_6} shows the RMSE performance as a function of SNR for LFM signals. It can be noted that, accuracy of all  methods except LDNN are improved. Although BFDNN does not explicitly perform DoA estimation, it achieves substantially lower RMSE than MUSIC, RDNN, and LDNN over all the evaluated SNR range. The Classification-DNN approach provides the lowest RMSE, while the proposed method exhibits competitive performance. The classification method demonstrates high RMSE at very low SNR level (below -30dB), where noise dominates and reduces feature separability. As the SNR increases beyond that, a sharp drop in RMSE is observed, indicating a transition
from noise-limited to signal-dominant conditions. This abrupt change reflects the thresholding behavior of classification models, wherein performance improves significantly once the input features become sufficiently distinguishable. At very low SNR around -30dB the proposed method provides lowest RMSE than all the methods compared. These results indicate that the beamforming solutions generated by BFDNN preserve useful directional information that can be exploited for target localization when required.

The algorithms that rely on grid search like MUSIC and classification- DNN, will impose a lower bound on estimation accuracy when the true angles lies off grid\cite{ref59}. This lower bound is dependent on the grid size. Therefore increase in SNR will not decrease RMSE, when the minimum error is reached in these cases \cite{ref60}.
In our proposed work, the angle is estimated from the radiation pattern by locating the peaks. So the minimum error to be obtained depends on the angular grid resolution. Here, the span $[-60^\circ, 60^\circ]$ is divided into 360 points. Consequently, the RMSE saturates at this grid-limited floor, which explains its flattening at high SNRs.
\subsection{Beam Pattern Comparison}
\begin{figure}[!t]
\centering
\includegraphics[width=3.25in,height=2.5in]{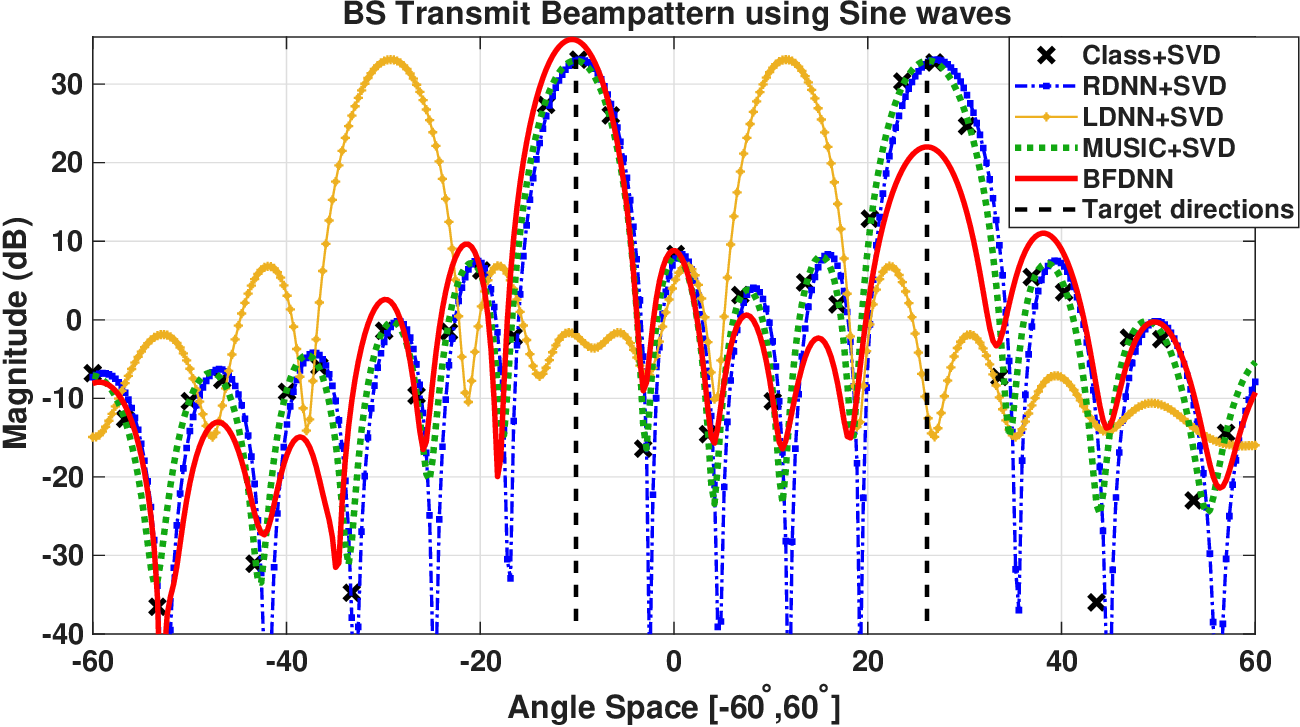}
\caption{Transmitter beam pattern for BS transmitting sine signals.}
\label{fig_7}
\end{figure}
\begin{figure}[!t]
\centering
\includegraphics[width=\columnwidth]{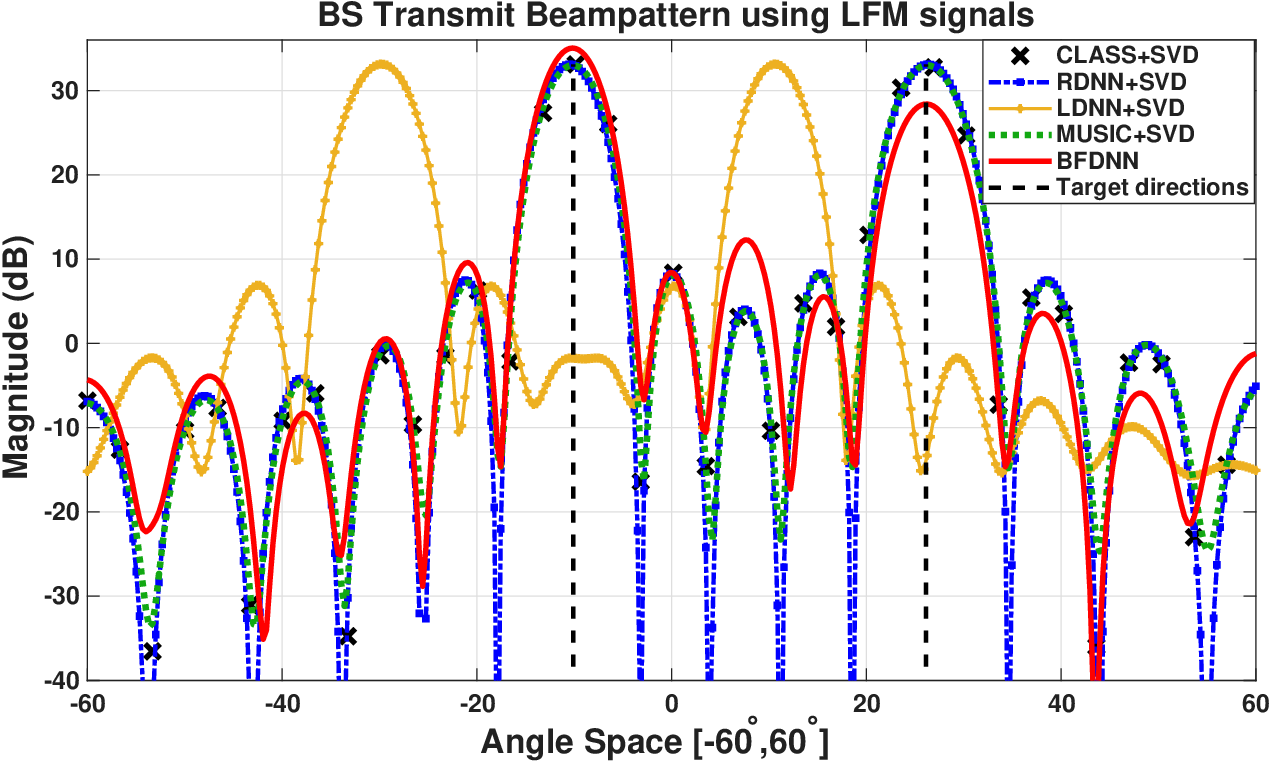}
\caption{Transmitter beam pattern for BS transmitting LFM signals.}
\label{fig_8}
\end{figure}

Based on the angles estimated using different methods beam forming weights are estimated using the SVD method mentioned in \cite{ref1}. The beampatterns obtained from those weights are plotted in Fig.\ref{fig_7} and Fig.\ref{fig_8} at 10 dB SNR for sine and LFM signals respectively. BFDNN directly predicts beamforming weights from covariance matrix. Eventhough in Fig. \ref{fig_7}, one of the mainlobes gain is less for BFDNN compared to other methods, in practical scenarios when using LFM signals radiation pattern obtained is almost as same as the other methods. As observed in Fig. \ref{fig_8}, the proposed method produces dominant main lobes near the desired target directions and exhibits beam patterns that are qualitatively similar to those obtained using the benchmark approaches. The LDNN based pipeline shows noticeable beam-pointing errors, while all other approaches generate beams in the intended directions. These results indicate that the proposed covariance-to-beamformer mapping captures sufficient spatial information to reproduce the desired beamforming behavior in the considered ISAC scenario.

The accurate beam placement achieved by BFDNN explains the strong indirect DoA-estimation performance observed in Fig. 4. Since target directions correspond to dominant beam peaks, the generated radiation pattern inherently contains directional information even though the network is trained only for beamforming-weight prediction.

\subsection{Effective Spectral Efficiency}

The performance of the downlink communication is evaluated by computing the  Effective Spectral Efficiency (ESE) for highly mobile scenarios \cite{ref61} as: 
\begin{align*} R_{\mathrm {eff}}=&~\left ({1-\frac {T_{\mathrm {tr}}}{T_{\mathrm {B}}}}\right) \times {R_{DL}} \tag{7}\end{align*}
where $T_{B}$ is the beam coherence time  which  represent the average time over which the beams stay aligned and first $T_{tr}$ time of every beam coherence time is allocated for the channel training/beamforming design. $R_{DL}$ is the downlink spectral efficiency (SE) \cite{ref1}.
\begin{figure*}[t]
\centering
\setlength{\tabcolsep}{1pt}

\subfigure[]{\includegraphics[width=0.32\textwidth,keepaspectratio]{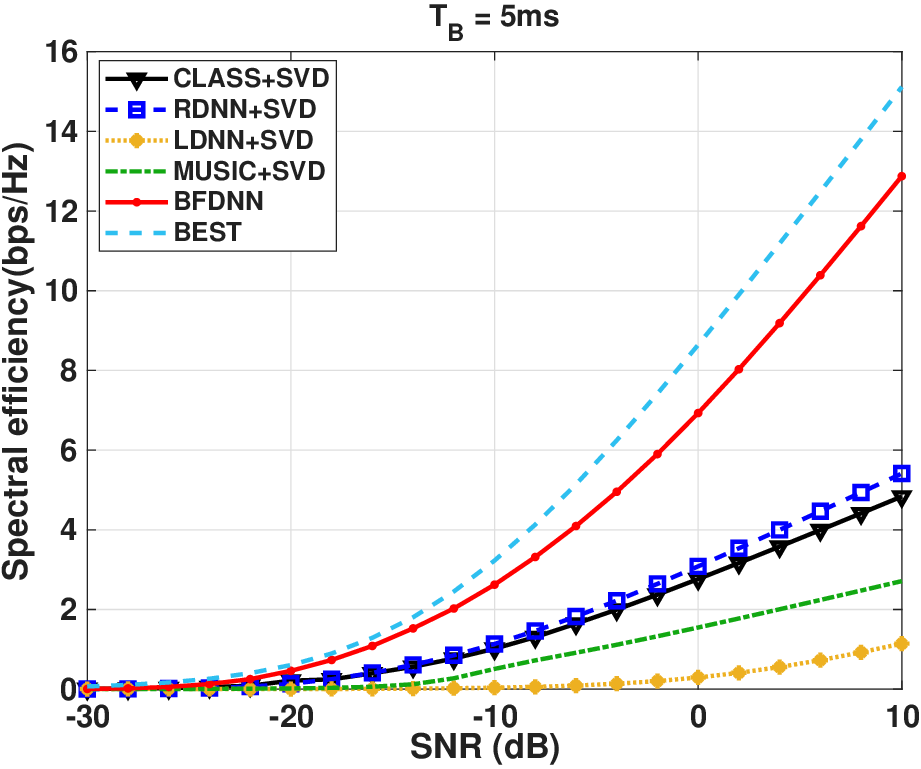}}
\subfigure[]{\includegraphics[width=0.32\textwidth,keepaspectratio]{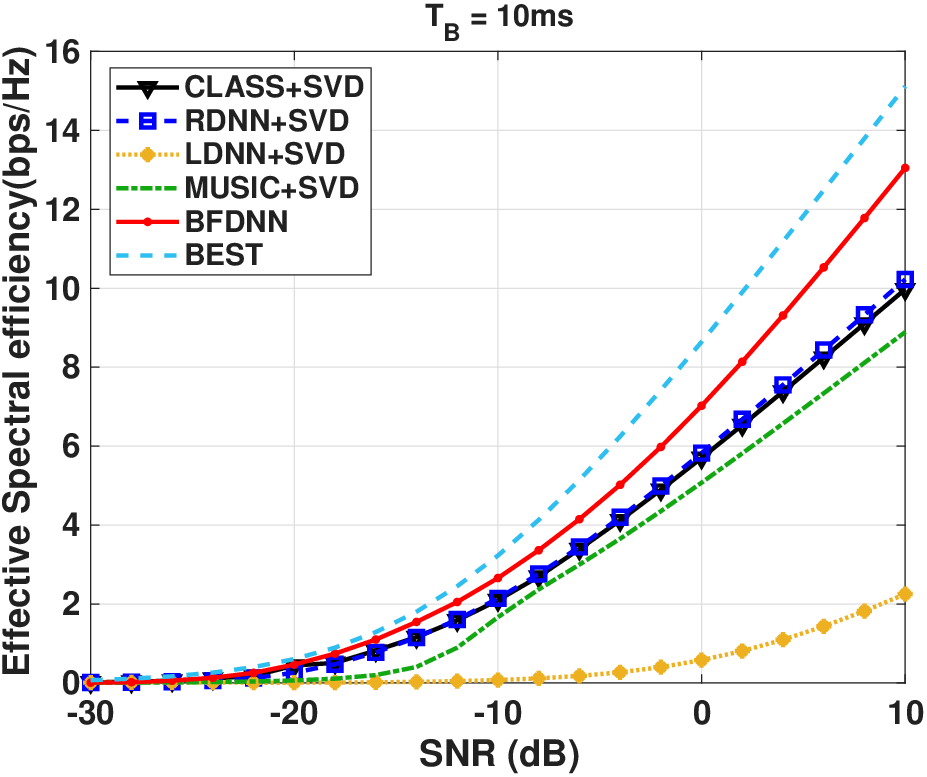}}
\subfigure[]{\includegraphics[width=0.32\textwidth,keepaspectratio]{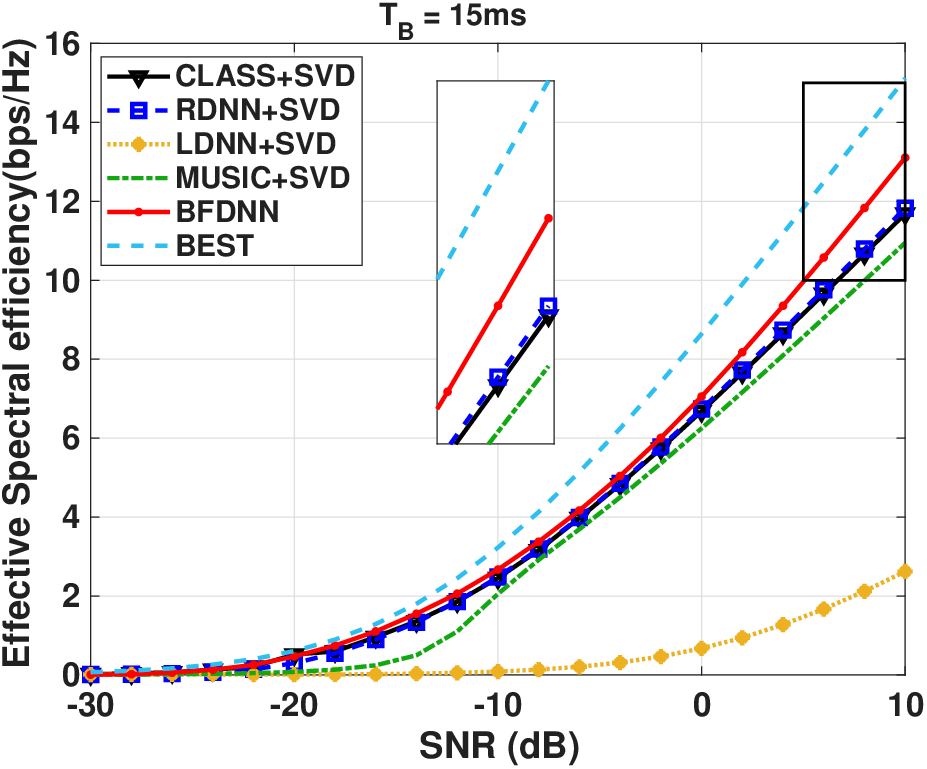}}
\caption{Effective spectral efficiency comparison of downlink communication under different beam coherence time for sine signals.}
\label{fig:se_all_SINE}
\end{figure*}

\begin{figure*}[t]
\centering
\setlength{\tabcolsep}{1pt}
\subfigure[]{\includegraphics[width=0.32\textwidth,keepaspectratio]{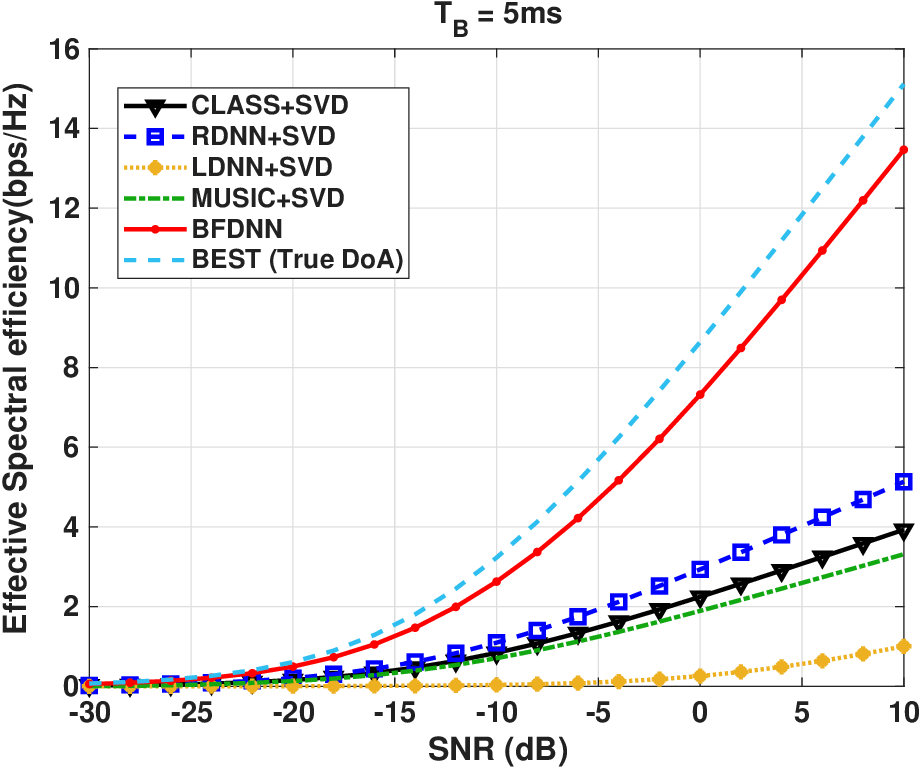}}
\subfigure[]{\includegraphics[width=0.32\textwidth,keepaspectratio]{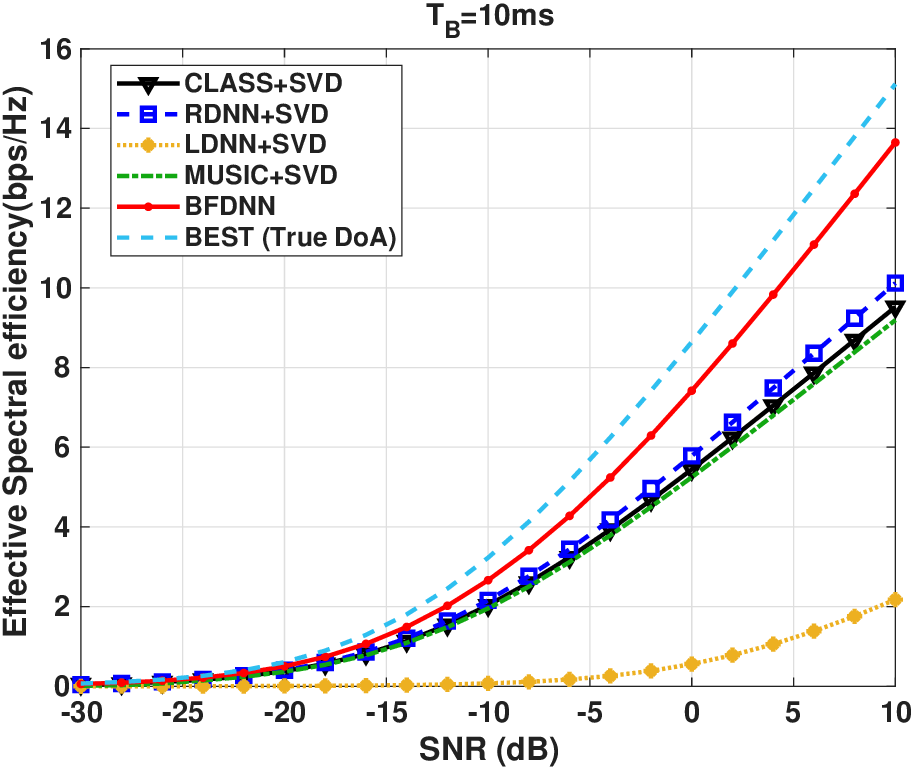}}
\subfigure[]
{\includegraphics[width=0.32\textwidth,keepaspectratio]{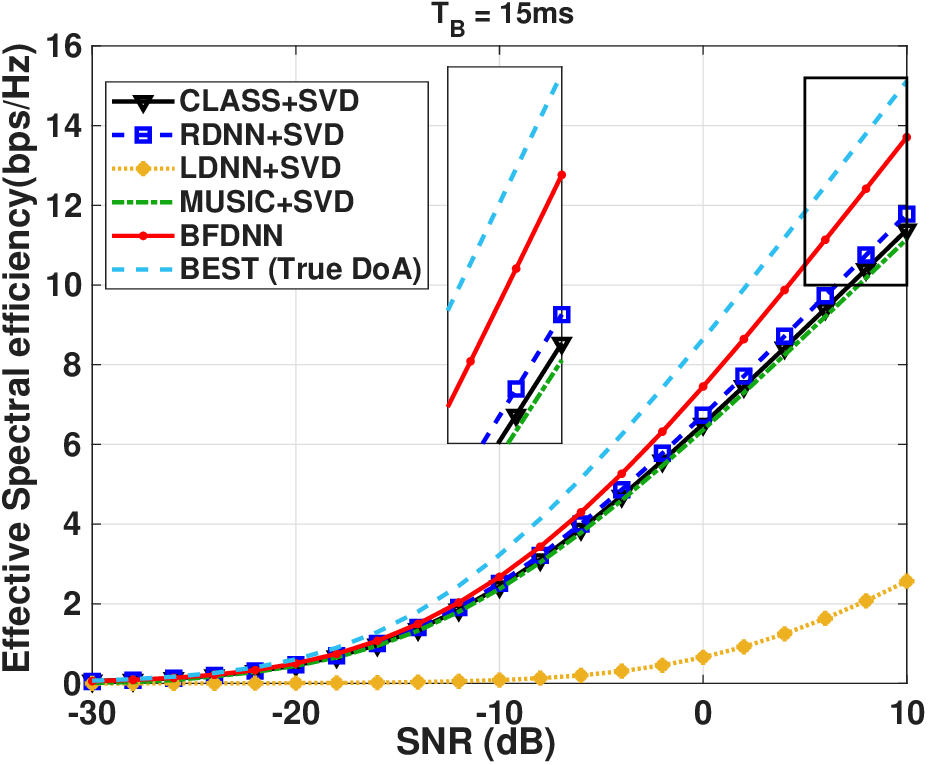}}

\caption{Effective spectral efficiency comparison of downlink communication under different beam coherence time for LFM.}
\label{fig:se_all}
\end{figure*}
Fig.\ref{fig:se_all_SINE} and Fig.\ref{fig:se_all} presents the ESE as a function of SNR for a representative beam coherence time. The spectral efficiency of all methods increases with SNR, as expected. The `BEST' spectral efficiency is plotted based on considering the true DoA and $T_{tr}$ as 0ms. The proposed BFDNN shows higher ESE compared to those of the benchmark approaches across the evaluated SNR range, for all $T_{B}=5$ms, $T_{B}=10$ms and $T_{B}=15$ms for both sine and LFM signals. ESE shows improvement for LFM signals compared to sine signals. At lower beam coherence time the ESE for the proposed method provides excellent performance compared to benchmark methods. These observations suggests that the beamforming solutions generated by BFDNN preserve the communication characteristics required for reliable ISAC operation, without introducing performance penalty when eliminating the intermediate DoA-estimation stage.
\subsection{Inference Latency Analysis}
The primary motivation of the proposed framework is the reduction of online beamforming latency. The performance of each method at 10dB SNR and $T_{B}=10$ms is summarized in Table.\ref{tab:table1} for LFM signals. The average prediction time required to generate beamforming weights for each method is provided in this table. The benchmark approaches require both DoA estimation and beamformer computation, resulting in prediction times between $3.3\times10^{-3}$ s and $3.9\times10^{-3}$s. In contrast, BFDNN directly generates beamforming weights through a single forward pass of the neural network and requires only $1.313\times10^{-4}$s. The measured results correspond to a latency reduction of approximately 25-30 times, relative to the benchmark methods on the evaluated platform.

Table~\ref{tab:table1} shows that the proposed BFDNN achieves competitive DoA RMSE and higher ESE at the lowest inference latency among all compared methods. These results indicate that direct covariance-to-beamformer learning  maintains competitive sensing accuracy and favorable communication performance in the considered ISAC scenario, demonstrating the efficiency of the proposed method.
\begin{table}[t]
\caption{Performance comparison
\label{tab:table1}}
\centering
\begin{tabular}{|c||c|c|c|}
\hline
\textbf{Methods} & \textbf{Prediction Time(s)} & \textbf{RMSE} & \textbf{ESE(bps/Hz)} \\
\hline
Classification+SVD & 0.0037 & \bf{0.1051} & 9.5204  \\
\hline
RDNN+SVD & 0.0033 &0.2808 & 10.1183\\
\hline
LDNN+SVD & 0.0035 & 17.7031 & 2.1756 \\
\hline
MUSIC+SVD & 0.0039 &  0.8011 & 9.1911\\
\hline
BFDNN & $\bf{1.313\times 10^{-4}}$ & \bf{0.1135} &\bf{13.6489} \\
\hline
\end{tabular}
\end{table}

\section{Conclusion}

In this letter, we proposed BFDNN, a covariance-to-beamformer learning framework for ISAC systems with reduced beamforming latency. Unlike the conventional two-stage pipeline that estimates the DoAs before generating the beamformer weights, the proposed framework directly predicts the hybrid beamforming weights from covariance-matrix  using a compact regression network, BFDNN. Simulation results show that the generated beam patterns are similar to those of the benchmark DoA-estimation-based approaches while achieving higher ESE in the considered ISAC scenario.   The proposed framework also enables indirect direction estimation from the generated beam pattern  with an RMSE lower than that of the MUSIC, RDNN, and LDNN based approaches and competitive with that of the classification-based method, which achieves the lowest RMSE. The main advantage of the framework is its reduced online latency: BFDNN lowers the beamforming prediction time from approximately $3.3\times10^{-3}$--$3.9\times10^{-3}$~s for the benchmark methods to $1.313\times10^{-4}$~s, a reduction of roughly 25--30 times on the evaluated platform. These results indicate that direct covariance-to-beamformer learning is a promising approach for low-latency beam management in future ISAC systems, particularly where rapid beam adaptation is required. Future work includes designing similar low-complexity architectures to extend the framework to mobility scenarios.

\vfill

\end{document}